\documentclass[journal,twoside,web]{ieeecolor}
\usepackage{tmi}
\usepackage{cite}
\usepackage{amsmath,amssymb,amsfonts}
\usepackage{algorithmic}
\usepackage{graphicx}
\usepackage{textcomp}

\usepackage[T1]{fontenc}
\usepackage{xspace}
\usepackage{amsmath}
\usepackage[titlenumbered,ruled]{algorithm2e}
\def\eg{\textit{e.g.}~}

\def\etal{\textit{et al.}\xspace}
\def\ie{\textit{i.e.,}~}

\def\vs{\emph{vs}.~}
 
\newcommand{\ra}[1]{\renewcommand{\arraystretch}{#1}}
\usepackage{color}

\newcommand{\JimSan}{Jim\'{e}nez-S\'{a}nchez}

\DeclareMathOperator{\EX}{\mathbb{E}}% expected value

\usepackage{booktabs}% http://ctan.org/pkg/booktabs
\usepackage{tabu}
\usepackage{multirow}
\usepackage{subfig}
\usepackage{comment}

\def\BibTeX{{\rm B\kern-.05em{\sc i\kern-.025em b}\kern-.08em
    T\kern-.1667em\lower.7ex\hbox{E}\kern-.125emX}}

\begin{document}

\title{Memory-aware curriculum federated learning \\ for breast cancer classification}
\author{Amelia \JimSan, Mickael Tardy, Miguel A. Gonz\'{a}lez Ballester, Diana Mateus, Gemma Piella
\thanks{Corresponding author: Amelia \JimSan~(amji@itu.dk). A. JS., G. P., and M. A. G. B. are with BCN MedTech, Department of Information and Communication Technologies, Universitat Pompeu Fabra, 08018 Barcelona, Spain. (e-mails: gemma.piella@upf.edu, ma.gonzalez@upf.edu). A. JS. is currently at IT University of Copenhagen (ITU), Copenhagen, Denmark. The revision
of this manuscript was done while A. JS. was at ITU. M. A. G. B. is also with ICREA, Barcelona, Spain.}
\thanks{M. T., and D. M. are with Ecole Centrale de Nantes, LS2N, UMR CNRS 6004, 44321, Nantes, France (e-mail: diana.mateus@ec-nantes.fr). M. T. is also with Hera-MI SAS (mickael.tardy@hera-mi.com).}
\thanks{D. M. and G. P. are joint senior authors.}
}

\maketitle

\begin{abstract}

\textbf{Background and Objective}: For early breast cancer detection, regular screening with mammography imaging is recommended. Routine examinations result in datasets with a predominant amount of negative samples. The limited representativeness of positive cases can be problematic for learning Computer-Aided Diagnosis (CAD) systems. Collecting data from multiple institutions is a potential solution to mitigate this problem. Recently, federated learning has emerged as an effective tool for collaborative learning. In this setting, local models perform computation on their private data to update the global model. The order and the frequency of local updates influence the final global model. In the context of federated adversarial learning to improve multi-site breast cancer classification, we investigate the role of the order in which samples are locally presented to the optimizers. 

\textbf{Methods:} We define a novel memory-aware curriculum learning method for the federated setting. We aim to improve the consistency of the local models penalizing inconsistent predictions, \ie forgotten samples. 
Our curriculum controls the order of the training samples prioritizing those that are forgotten after the deployment of the global model. Our approach is combined with unsupervised domain adaptation to deal with domain shift while preserving data privacy.

\textbf{Results:} Two classification metrics: area under the receiver operating characteristic curve (ROC-AUC) and area under the curve for the precision-recall curve (PR-AUC) are used to evaluate the performance of the proposed method. Our method is evaluated with three clinical datasets from different vendors. An ablation study showed the improvement of each component of our method. The AUC and PR-AUC are improved on average by 5\% and 6\%, respectively, compared to the conventional federated setting. 

\textbf{Conclusions:} We demonstrated the benefits of curriculum learning for the first time in a federated setting. Our results verified the effectiveness of the memory-aware curriculum federated learning for the multi-site breast cancer classification. Our code is publicly available at: \mbox{https://github.com/ameliajimenez/curriculum-federated-learning.}
\end{abstract}


\vspace{2em}

\section{Introduction}
\label{sec:introduction}
\IEEEPARstart{B}{reast} cancer is the most commonly occurring type of cancer worldwide for women \cite{Sung2021}. Early detection and diagnosis of breast cancer is essential to decrease its associated mortality rate. The medical community recommends regular screening with X-ray mammography imaging for its early detection and follow-up. High-resolution images showing tissue details need to be analyzed to spot abnormalities and to provide a precise diagnosis. For instance, suspicious micro-calcifications can be smaller than $0.5~{\rm mm}$ \cite{Tse2008, Mercado2014}. Despite high incidence (\ie 12\%) \cite{Sung2021}, the extensive breast cancer screening results predominantly in negative samples. The limited representativeness of positive cases can be problematic for learning-based Computer-Aided Diagnosis (CAD) systems. A potential solution to mitigate this problem and to increase the size of the annotated dataset is to employ data coming from multiple institutions. However, sharing medical information across (international) institutions is challenging in terms of privacy, technical and legal issues. Secure and privacy-preserving machine learning offers an opportunity to bring closer patient data protection and data usage for research and clinical routine purposes. 

%%%%%%%%%%%%%%%%%%%%%%%%%%%%%%%%%%%%%%%%%%%%%%%%%%%%% 
\begin{figure*}[t]
    \centering
    \includegraphics[width=0.85\textwidth]{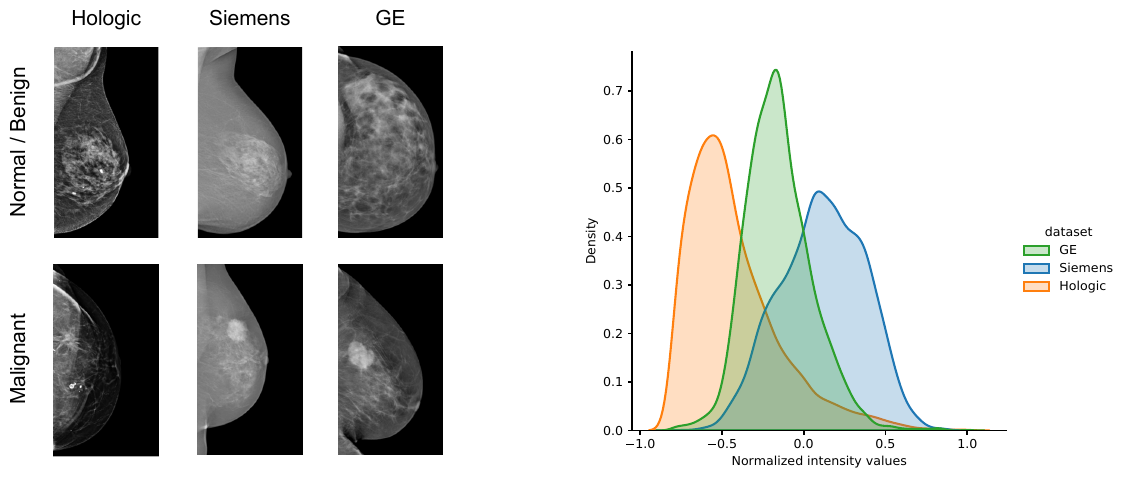}%
    \caption{Left: Exemplary mammograms of benign and malignant cases. Right: pixel-intensity distributions of different sites.} %
    \label{fig:domain-shift} %
\end{figure*}
%%%%%%%%%%%%%%%%%%%%%%%%%%%%%%%%%%%%%%%%%%%%%%%%%%%%%%%%%%%%%%%%%%%%%%

Federated Learning (FL) aims to train a machine learning algorithm across multiple decentralized nodes holding locally the data samples, \ie without exchanging them. Training such a decentralized model in a FL setup presents three main challenges~\cite{Darzidehkalani2022ReviewFederated}: (i) system and statistical heterogeneity, (ii) data protection, and (iii) distributed optimization. We deal with the three challenges for breast cancer classification in the context of FL. 

The first challenge concerns system and data heterogeneity. For the same imaging modality, different system vendors produce images following significantly different intensity profiles. To cope with such diversity, Zhao~\etal~\cite{Zhao2018FedNonIID} have proposed to create a small subset of data that is globally shared between all sites for image classification and speech recognition, and recent works \cite{Peng2019FedDA,Li2021FedBN} have integrated Unsupervised Domain Adaptation (UDA) into the FL framework. UDA methods force the model to learn domain-agnostic features through adversarial learning \cite{Peng2019FedDA} or a specific type of batch normalization \cite{Li2021FedBN}. In this work, we do not share image data between the sites and follow an UDA adversarial approach to handle domain shift.

To address the second challenge, data protection, cryptographic techniques \cite{Bonawitz2017Encrypt} or differential privacy \cite{Dwork2006DP,Dwork2014DP} are employed. Differential privacy perturbs the parameters of each local model by purposely adding noise before uploading them to the server for aggregation. We leverage differential privacy for data protection in our method.

The third challenge concerns the distributed optimization in the FL setting. Individual models are trained locally on private data and the central server is responsible for the global aggregation of the local updates. Usually, the communication of the local models to the server occurs a certain number of times every epoch. Therefore, we propose a novel curriculum learning approach that provides a meaningful order to the samples.

\paragraph*{\textbf{Contributions}} 
\begin{itemize}
\item In this work, we investigate for the first time the use of Curriculum Learning (CL)~\cite{Bengio2009CL} to improve multi-site breast cancer classification in a federated setting. 
\item Our CL approach is implemented via a data scheduler~\cite{Jimenez2020CLPriorUncertainty}, which establishes a prioritization of the local training samples. In this work, we design the schedulers to improve the consistency of the local models penalizing inconsistent predictions, \ie forgotten samples. 
\item We combine our data scheduler with federated adversarial learning and show that the association is beneficial for the classification and the alignment between domain pairs. 
\item Our CL strategy is shown effective for the multi-site breast cancer classification on high-resolution mammograms of clinical datasets from different vendors.
\end{itemize}

%%%%%%%%%%%%%%%%%%%%%%%%%%%%%%%%%%%%%%%%%%%%%%%%%%%%% 
\begin{figure*}[t]
    \centering
    \includegraphics[width=0.9\textwidth]{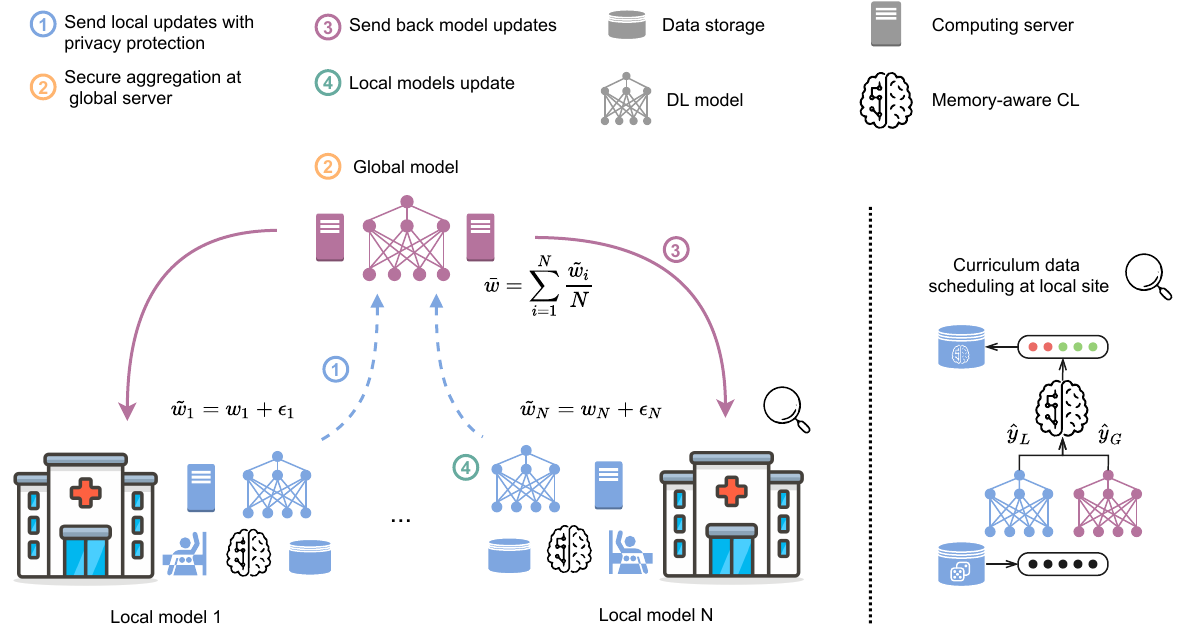}%
    \caption{Memory-aware curriculum federated learning framework with data privacy protection. (1) Each local model shares its weights after the addition of Gaussian noise (dotted blue arrows). (2) The global server performs the aggregation of the local models' weights. (3) The resulting averaged model is deployed to each site (purple arrows). (4) Local models are updated. The curriculum data scheduler rearranges the training samples to prioritize samples that were forgotten after the deployment of the global model.}%
    \label{fig:fedlearning} %
\end{figure*}
%%%%%%%%%%%%%%%%%%%%%%%%%%%%%%%%%%%%%%%%%%%%%%%%%%%%%%%%%%%%%%%%%%%%%%

\section{Related work} \label{sec:related}

\subsection{Federated Learning}  
FL arises from the need of sharing sensitive medical data between different healthcare providers. FL has been mainly formulated in two ways: (i) differential privacy \cite{Dwork2006DP,Dwork2014DP}, \ie each site trains a local model with private data and only shares model parameters \cite{Zhao2018DifferentialPrivacy}, and (ii) protecting the details of the data using cryptographic techniques \cite{Bonawitz2017Encrypt}, such as secure multi-party computation \cite{Mohassel2018Secure} and homomorphic encryption \cite{Hardy2017Homomorphic}. We focus on the differential privacy approach.

Only few FL works have been shown effective on medical images. For instance, for prediction of disease incidence, patient response to treatment, and other healthcare events~\cite{Huang2019FLEHR}; for leveraging decentralized unlabeled data~\cite{Dong2021FCL}; for Magnetic Resonance Imaging (MRI) harmonization across studies~\cite{Dinsdale2022FedHarmony}. FL approaches have been leveraged for applications such as brain tumor segmentation \cite{Sheller2018MultiSegmentation, Li2019PrivacySegmentation, Bercea2021feddis}; survival prediction on histopathology whole slide images~\cite{Lu2022WSI}; and lately for classification~\cite{Grimberg2020WeightErosion, Andreux2020SiloedFL, Li2020FLDA, Wang2020CrossSiteCOVID, Yeganeh2020IDA}, \cite{Luo2022FedSLD}. To the best of our knowledge there is only one prior FL work in the context of breast imaging \cite{Roth2020Federated} dealing with the problem of density classification. Different to Roth~\etal~\cite{Roth2020Federated} we address the more challenging malignancy classification task, which requires dealing with higher resolution images classification as input. We also integrate in our model a solution for  the domain shift problem, while proposing a new optimization strategy to update the local and global models' weights.

\subsection{Domain Adaptation} 
Deep learning methods assume that samples from the training (source) and testing (target) set are IID data. However, this statement does not always hold. When the data distribution from the source and target domains is related but different, there is a domain shift. Domain Adaptation (DA) aims to remove such shifts by transferring the learned representation from a source to a target domain. When target labels are unavailable during the training phase, UDA techniques are employed. One of the UDA strategies is to learn a domain-invariant feature extractor, which aligns the feature distribution of the target domain to that of the source by: (i) minimizing a distance of domain discrepancy \cite{Long2013JointDistribAdapt}, (ii) revisiting batch normalization layers \cite{Carlucci2017BatchDA}, or (iii) through adversarial learning \cite{Ganin2016DANN}. 

Despite less annotation requirements, the above UDA approaches need access to both source and target data \cite{long2015learning, Ganin2015UDA}. However, in the federated setting, data is stored locally and cannot be shared. Recently, federated batch normalization \cite{Li2021FedBN} and federated adversarial domain adaptation \cite{Peng2019FedDA, Peterson2019FedDAPrivate} have been proposed to deal with DA under the privacy-preserving requirement. The work by Li~\etal~\cite{Li2021FedBN} focus on mitigating \textit{feature shift}, \ie the deviation in feature space, using batch normalization before averaging the local models. In contrast, Peng~\etal~\cite{Peng2019FedDA} train in an adversarial manner a feature extractor and a domain discriminator to learn a domain-invariant representation and alleviate domain shift. The latter has been applied to functional MRI (f-MRI) on 1-D signal data using a multi-layer perceptron \cite{Li2020FLDA}. We extend this approach to deal with high-resolution images and to interact with our proposed CL method.

\subsection{Curriculum Learning}  
CL~\cite{Bengio2009CL} is inspired in the \textit{starting small} concept from cognitive science. CL methods follow a systematic and gradual way of learning. A scoring function is defined to determine the priority of the training samples. Based on this scoring function, which can measure, for example, difficulty or uncertainty, the training samples are weighted or presented in a certain order to the optimizer. This new order has an impact on the local minimum achieved by the optimizer, leading to an improvement in the classification accuracy.

CL has already demonstrated an improved performance in medical image classification tasks, such as thoracic disease \cite{Tang2018AttentionGuidedCL}, skin disease \cite{Yang2019SPBLSkin}, proximal femur fractures \cite{Jimenez2019MedicalCurriculum, Jimenez2020CLPriorUncertainty} and breast screening classification \cite{Maicas2018likeRadiologists}. These techniques exploit either attention mechanisms \cite{Tang2018AttentionGuidedCL}, meta-learning \cite{Maicas2018likeRadiologists}, prior knowledge \cite{Jimenez2019MedicalCurriculum, Yang2019SPBLSkin} or uncertainty in the model's predictions \cite{Jimenez2020CLPriorUncertainty}. 

There is little prior work in CL in combination with DA techniques for general classification. Mancini~\etal~\cite{Mancini2020CuMix} investigated a combination of CL and Mixup \cite{Zhang2017mixup} for recognizing unseen visual concepts in unseen domains. Shu~\etal~\cite{Shu2019TransferableCL} addressed two entangled challenges of weakly-supervised DA: sample noise of the source domain, and distribution shift across domains. An extreme case of DA is that of zero-shot learning, in which at test time, a learner observes samples from classes that were not observed during training. Tang~\etal~\cite{Tang2020CMSS} proposed an adversarial agent, referred to as curriculum manager, which learns a dynamic curriculum for source samples.

Different from \cite{Mancini2020CuMix, Zhang2017mixup, Shu2019TransferableCL, Tang2020CMSS}, which aim at improving transferability between domains, we choose to schedule the data within each domain. The scoring function of our data scheduler is inspired by temporal ensembling~\cite{Laine2017TemporalEnsembling} and consistency training~\cite{Xie2020ConsistencyTraining} in semi-supervised learning. In temporal ensembling a consensus of predictions is formed for unlabeled samples using the outputs of the network-in-training at different epochs. In consistency training, model predictions are constrained to be invariant to input noise. We design local data schedulers aiming to deliver consistent predictions between the global and the local models. To prevent forgetting samples that were previously correctly classified by the local model, we monitor the training samples before and after the deployment of the global model. We define a scoring function that assigns high values to samples that have been forgotten by the local model. Thus, our CL method builds locally memory-aware data schedulers to improve model consistency.
%%%

\section{Methods} \label{sec:method}
In this section, we formulate the details of our proposed curriculum approach to locally schedule training samples in the FL setting. The overall FL framework is depicted in Fig.~\ref{fig:fedlearning}. In this setting, we train a local model per site. We assume that each site has local data storage, a computing server, and a memory-aware CL module. Nevertheless, at the global level, no imaging data are stored. We consider a computing node and space to store only the global model. In this type of FL setting, both global and local models share the same architecture, and it is common to share the model weights and aggregate them at the central server. Moreover, local healthcare providers may have diverse imaging systems resulting in datasets with different intensity profiles. To ease the existing domain shift between the sites, we deploy an UDA strategy that shares the latent representations (and not the image data) between domain pairs. Both the model weights and the embeddings are blurred with Gaussian noise \cite{Li2020FLDA} to protect the private data using differential privacy \cite{Dwork2006DP, Dwork2014DP}. Our memory-aware CL module compares the local and global model predictions and assigns scores to each training sample. The data scheduler leverages the curriculum probabilities to locally arrange the samples.

In Subsection~\ref{subsec:multi-site}, the overall FL framework is presented. In Subsection~\ref{subsec:fed}, we present the details of the FL setup with data privacy-preserving scheme. Then, in Subsection~\ref{subsec:fed-align}, we introduce DA into the framework. And finally, in Subsection~\ref{subsec:fed-align-cm} we present the details of our proposed method leveraging CL to avoid forgetting locally learned samples in the FL setting.

\subsection{Multi-site learning} \label{subsec:multi-site}
Next, we develop our method to learn a collaborative CAD system in a decentralized multi-site scenario with a privacy-preserving strategy. Let us denote each site's dataset as $\mathcal{D}_n$ where $n = 1,...,N$ and $N$ is the total number of sites. Each dataset is composed of mammography images $X_n$ and their corresponding diagnosis $Y_n$, \ie $\mathcal{D}_{n}=\{X_n,Y_n\}$. We aim to detect malignant cases by training a deep-learning model. We formulate the learning objective as a binary classification task, where malignant samples correspond to the positive class. Each local model aims to minimize the cross-entropy loss over the training data from a particular site $n$:
\begin{align} \label{eq:lossCls}
    \mathcal{L}_{Cls, n} = - \sum_{n_k} y_{n_k} \log(p_{n_k}) + (1-y_{n_k}) \log(1-p_{n_k}),
\end{align}
where $y_{n_k}$ is the label of the $k$-th subject in the training label set ${Y_n=\{y_{n_1},...y_{n_{|Y_n|}}\}}$ and $p_{n_k}$ is the corresponding output probability of the model for an input $x_{n_k} \in X_n$. As depicted in Fig.~\ref{fig:localmodels}-\textit{left}, we split the deep learning model into a feature extractor $F$ and a classifier $Cls$. We refer to the output of the feature extractor as the latent representation or embedding. In this work, we assume the most challenging scenario, in which we consider that each site has mammography systems of different vendors (see Fig.~\ref{fig:domain-shift}).

\subsection{Federated learning} \label{subsec:fed} 
We assume that data owners collaboratively train a global model without sharing their image data. The term \textit{federated} was coined because the learning task is solved by a federation of participating models (frequently referred to as \textit{clients}), which are coordinated by a central \textit{server}.

The FL scenario is depicted in Fig.~\ref{fig:fedlearning}. We assume that each local site has data storage and a computing node. Nevertheless, at the global level, only computing is possible. Once that individual models have been trained on private data, there are four key steps in the FL training process: (1) local updates are sent to the global server with privacy protection or encryption, (2) the central server aggregates the local updates, (3) the aggregated model parameters are deployed to the local sites, and (4) local models are updated. After that, a new round of local training starts.

To apply SGD in the federated setting, each client $n$ computes gradients on the full local data for the current model, and the central server performs the aggregation of these weights to build a global update. Let us assume a fixed learning rate $\eta$ and denote the gradients at each client as $g_n$. The central server computes the update as $w_{t+1} \xleftarrow[]{} w_t - \eta \sum_{n=1}^{N}\frac{m_n}{M} g_n$, where $m_n$ is the number of images at site $n$, and $M$ the total number of images. We refer to this algorithm as \texttt{FedSGD}. We can decompose the global update into local client ones: first, one takes a gradient descent step from the current model using each local dataset, $\forall n\, w_{t+1}^{n} \xleftarrow{} w_t - \eta\, g_n$. Then, we let the server make a weighted average of the resulting local updates as $w_{t+1} \xleftarrow[]{} \sum_{n=1}^{N} \frac{m_n}{M}w_{t+1}^n$. Instead of performing one global update after each local computation, we can add multiple iterations of the local update to each client before the averaging step. Global model updates are then performed at every \textit{communication round}. Let us denote: $Q$, the total number of optimization iterations; $\tau$, the communication pace; and $B$, the local mini-batch size used for the client updates. In each epoch, the communication between the models happens $Q/\tau$ times. Federated averaging \texttt{FedAvg} is a generalization of \texttt{FedSGD}~\cite{McMahan2017FedAvg}, which allows local nodes to perform more than one batch update on local data and exchanges the updated weights rather than the gradients. We build on top of \texttt{FedAvg} to further consider domain alignment. 

% Federated learning with domain alignment
\subsection{Federated adversarial learning} \label{subsec:fed-align} 
\begin{figure}[t]
    \centering
    \includegraphics[width=0.49\textwidth]{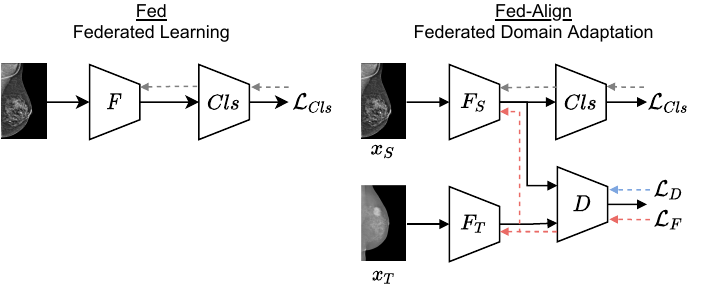}%
    \caption{Architecture comparison of \textit{left}: Fed and \textit{right}: Fed-Align. Colour dotted lines indicate backward passes with respect to each loss function. ($\mathcal{L}_{Cls}$: Eq.~\eqref{eq:lossCls}, $\mathcal{L}_{D}$: Eq.~\eqref{eq:lossD}, $\mathcal{L}_{F}$: Eq.~\eqref{eq:lossF}, $F$: feature extractor, $Cls$: classifier, $D$: domain discriminator).} %
    \label{fig:localmodels} %
\end{figure}

Medical images collected from different healthcare providers may originate from diverse devices or imaging protocols, leading to a domain shift problem when deploying algorithms trained on a single domain. In this scenario, we try to compensate the domain shift between every pair of domains. There is extensive literature on UDA methods \cite{Ganin2015UDA, Hoffman2018MultiSourceAdaptation, Zhu2017UnpairedImagetoImage}. However, these works do not generally satisfy the conditions of a FL setting: namely, that data should be stored locally and not shared. To satisfy the requirements of the FL framework and to address the domain shift problem, we adapt federated adversarial alignment \cite{Peng2019FedDA}. This method aligns the feature space by progressively reducing the domain shift between every pair of sites. To preserve privacy, only the noisy latent representations (Gaussian noise is added to each local latent representation) are shared between the sites at every communication round. This method leverages a domain-specific local feature extractor $F$, and a global discriminator $D$. For source $\mathcal{D}^S$ and target $\mathcal{D}^T$ sites, we train individual local feature extractors $F^S$ and $F^T$, respectively. For each site, we train a domain discriminator $D$ to align the distributions, see Figure~\ref{fig:localmodels}. While the feature extractors and the classifiers are combined, this is not the case for the discriminators, which remain independent for each site.

Optimization takes place in two iterative steps. In the first, the objective for discriminating the source domain from the others is defined as:
%Adversarial domain loss:
\begin{align} \label{eq:lossD}
\begin{split}
    \mathcal{L}_D(X^S, X^T, F^S, F^T) = 
    & -\EX_{x^S \sim X^S} [\text{log}\, D\,(F^S(x^S))] \\
    & - \EX_{x^T \sim X^T} [\text{log}\,(1-D\,(Z \circ F^T(x^T))],
\end{split}
\end{align}
\noindent where $Z(\cdot)$ is the Gaussian noise generator for privacy preservation, see Algorithm~\ref{alg:framework} (Supplementary Material) for further details. In the second step, we consider the adversarial feature extractor loss:
\begin{align} \label{eq:lossF}
\begin{split}
    \mathcal{L}_F(X^S, X^T, D) = 
    & -\EX_{x^S \sim X^S} [\text{log}\, D\,(F^S(x^S))] \\
    & - \EX_{x^T \sim X^T} [\text{log}(D\,(Z \circ F^T(x^T))].
\end{split}
\end{align}
The weights of the feature extractors $F^S$ and $F^T$ and the domain discriminator $D$ remain unchanged for the first and second step, respectively. 

%%%%%%%%%%%%%%%%%%%%%%%%%%%%%%%%%%%%%%%%%%%%%%%%%%%%% 
%%%%%%%%%%%%%%%%%%%%%%%%%%%%%%%%%%%%%%%%%%%%%%%%%%%%%%%%%%%%%%%%%%%%%%

% Scheduling source/training data in the Federated Setting
\subsection{Memory-aware curriculum federated learning} \label{subsec:fed-align-cm}
We propose to incorporate CL to improve the classification performance of the federated adversarial learning approach. In particular, the implementation of the curriculum is in the form of a data scheduler. A data scheduler is a mechanism that controls the order and pace of the training samples presented to the optimizer. Building from our previous work \cite{Jimenez2020CLPriorUncertainty} we design a CL approach adapted to the federated setting. In the following, we introduce the required components to define our CL method. We formalize the definition of the data scheduler through three components: a scoring function $\rho$, curriculum probabilities $\gamma$, and a permutation function $\pi$, and provide further details in the next paragraph.

The key element of our approach is the scoring function $\rho$, which is specific for FL. The scoring function $\rho$ assigns a score to every sample, which normalized becomes a curriculum probability $\gamma$. 
These probabilities are then used to sample the training set $\{X,Y\}$: the higher the probability, the more likely the sample is earlier presented to the optimizer. The sampling operation establishes a permutation $\pi$ determining the reordered dataset $\{X_{\pi},Y_{\pi}\}$, finally fed in mini-batches to the optimizer.

We consider a dynamic approach in which the scoring values are computed at every epoch $e$ for every training sample $k$. We get the predictions at every site $n$, before and after the communication between the models, obtaining local and global predictions $\hat{y}_L$, $\hat{y}_G$, respectively. To improve model consistency in the FL setting, our scoring function $\rho$ assigns higher values (thus higher curriculum probabilities $\gamma$) to samples that were forgotten. We consider that a sample was forgotten when its prediction changed to incorrect after the global model was deployed. 
The order in which samples are presented to the optimizer is determined by the curriculum probabilities $\gamma$. Our function is defined as:
\begin{equation} \label{eq:clweights}
  \rho_k^{(e)} =
  \begin{cases}
    2.0 & \text{if $\hat{y}_{L,k}^{(e)} = y_{k}$ and $ \hat{y}_{G,k}^{(e)} \neq y_{k}$} \\
    1.0 & \text{otherwise.} \\
  \end{cases}
\end{equation}

We propose a simple but efficient weighting. We emphasize learning of samples for which the prediction changed from correct to wrong after the model aggregation. Our scoring function is related to temporal ensembling~\cite{Laine2017TemporalEnsembling} and consistency training~\cite{Xie2020ConsistencyTraining}, which are used in semi-supervised setups. In a similar manner, we aim to improve the consistency of the local models penalizing inconsistent predictions, \ie forgotten samples. As shown in Eq.~\eqref{eq:clweights}
we compare the predictions before and after model aggregation, thus we refer to our approach as memory-aware. Our memory-aware curriculum federated learning method is summarized in Algorithm~\ref{alg:framework} (Supplementary Material).

\section{Experimental validation} \label{sec:experimental}
In order to validate the effect of data scheduling on the breast cancer classification in a federated setting with heterogeneous data. We perform experiments with two private and one public dataset. We compare our proposed approach combining FL, DA and CL, against FL alone and FL with DA.

\subsection{Datasets}
For our study, we employ 3 datasets of Full Field Digital Mammography (FFDM), coming from three different vendors: Hologic, GE and Siemens (INBreast~\cite{Moreira2012INBreast}). The first two are private clinical datasets, and the last one is publicly available. Institutional board approvals were obtained for each of the datasets. Intensity profiles \cite{Shen2019} among the datasets varied significantly, as can be observed in Fig.~\ref{fig:domain-shift}. This variability is mainly due to the different mammography systems and acquisition protocols used to generate digital mammograms. 
We do not use any site-specific image filtering to compensate the domain shift, and we apply the same preprocessing to the images from the three sites. The preprocessing consists of standard normalization with mean subtraction and division by the standard deviation. Each dataset was split into three parts with the ratio approximately of 70\%:10\%:20\% to build respectively the training, validation and test sets. Our problem is formulated as a binary classification task. The number of samples per class and database can be found in Table~\ref{table:datasets}. The negative class reunites benign findings and normal cases, the positive class contains only malignant cases confirmed with a biopsy. Mammography images are of different sizes, we cropped the empty rows and columns, and resized to 2048 pixels in height, and then padded to 2048 pixels in width. It is often the case that important cues for diagnosis are subtle findings in the image, which could be as small as 10 pixels in length \cite{Mercado2014}. Therefore, we do not apply any further downsampling and use a resolution of 2048 pixels, close to the original one.

%%%%%%%%%%%%%%%%%%%%%%%%%%%%%%%%%%%%%%%%%%%%%%%%%%%%%%%%%%%%%%%%%%%%%%
%% Table: Summary dataset
\begin{table}[t]  
\centering
\ra{1.5} 
\caption{Summary of the datasets used in this study.}
\begin{tabular}{@{}l|ccc@{}} \toprule
& Hologic & Siemens & GE \\ \midrule
Total Subjects & 1460 & 410 & 852 \\ 
Benign/Normal & 730 & 287 & 421 \\
Malignant & 730 & 123 & 431 \\
\bottomrule
\end{tabular}
\label{table:datasets}
\end{table}
%%%%%%%%%%%%%%%%%%%%%%%%%%%%%%%%%%%%%%%%%%%%%%%%%%%%%%%%%%%%%%%%%%%%%%

\subsection{Experimental Setting}
We perform an in-depth evaluation of our proposed method \textit{Fed-Align-CL} with a series of experiments. First, we investigate the effect of different pretraining strategies in the FL framework. Second, we compare the classification performance of our approach against other non-federated and federated approaches. Third, we investigate the influence of DA and CL in the resulting feature embeddings of the different methods.

\subsection{Implementation details} \label{subsec:impl-details}
\paragraph*{\textbf{Architectures}}
We employ as feature extractor $F$ the architecture proposed by Wu~\etal~\cite{Wu2019Breast}, a ResNet-22~\cite{He2015ResNet} that is adapted to take high-resolution images ($\sim 4$ megapixels) as input, and achieved state-of-the-art classification performance for breast cancer screening. We initialize the feature extractor with the pretrained weights provided by Wu~\etal~\cite{Wu2019Breast}\footnote[1]{https://github.com/nyukat/breast\_cancer\_classifier}. The weights of the classifier $Cls$ and domain discriminator $D$ are randomly initialized. The classifier $Cls$ is formed by 3 fully-connected layers. The first two are followed by batch normalization, ReLu activation, and dropout. The architecture for the domain discriminator $D$ is formed by two fully-connected layers with a ReLu activation in between and a sigmoid layer for the final output. Details of the architecture of the models can be found in Table~\ref{table:architecture} of the Supplementary Material. Our memory-aware CL approach builds on top of the federated adversarial learning code provided by Li~\etal~\cite{Li2020FLDA}\footnote[2]{https://github.com/xxlya/Fed\_ABIDE}. Different from \cite{Li2020FLDA} that employs a multi-layer perceptron for 1-D f-MRI signals, we deploy a specific CNN for high-resolution mammography images. The models were implemented with PyTorch~\cite{Paszke2019Pytorch} and ran on an Nvidia Titan XP GPU.

\paragraph*{\textbf{Hyperparameters}}
We train our models 5 times with different seed initialization for the classifier $Cls$ and domain discriminator $D$. Adam optimization is used for 50 epochs with an initial learning rate of $1\mathrm{e}{-5}$. We compute the adversarial domain loss $\mathcal{L}_D$, and also introduce the curriculum data scheduling, after training the feature extractor $F$ and classifier $Cls$ for 5 epochs. The dropout rate for the classifier $Cls$ is set to 0.5. The number of optimization iterations $Q=120$, and the local batch size $B_n=\lfloor m_n/Q \rfloor$. In each epoch, local models are updated according to the communication pace $\tau$. The shared weights are modified by the addition of random noise $\epsilon$ to protect data from inverse interpretation leakage. We generated Gaussian noise $\epsilon \sim N(0,s_{h}^{2} \sigma^2)$, assuming a sensitivity $s_{h}=1$ and a variance $\sigma^2=0.001$. We investigated different communication paces $\tau=\{10,20,40,60\}$, and noise values $\sigma^2=\{0,0.001,0.01,0.1\}$. We did not find significant differences in classification accuracy for the different communication paces $\tau$. There is a direct correlation between the amount of noise introduced in the system and the model performance, we consider that adding a noise $\sigma^2 = 0.001$ is a good trade-off.

\paragraph*{\textbf{Evaluation metrics}}
For the classification task, we report the area under the receiver operating characteristic curve (ROC-AUC) and AUC for the precision-recall curve (PR-AUC).

%%%%%%%%%%%%%%%%%%%%%%%%%%%%%%%%%%%%%%%%%%%%%%%%%%%%%%%%%%%%%%%%%%%%%%
%% Table: FL initialization of local models
\begin{table}[t] 
\centering
\ra{1.5} 
\caption{AUC of the federated learning method using different initialization approaches.}
\begin{tabular}{@{}l|c|c|c|c@{}} \toprule
 Initialization $\setminus$ Site & \textbf{Hologic} & \textbf{Siemens} & \textbf{GE} & \textbf{AVG} \\ \midrule
Local model & 0.57 & 0.38 & 0.66 & 0.53 \\
Scratch & 0.73 & 0.52 & 0.65 & 0.63 \\
DDSM & 0.69 & 0.62 & 0.65 & 0.65 \\
Wu~\etal~\cite{Wu2019Breast} & \textbf{0.78} & \textbf{0.65} & \textbf{0.83} & \textbf{0.75} \\
\bottomrule
\end{tabular}
\label{table:initialization}
\end{table}
%%%%%%%%%%%%%%%%%%%%%%%%%%%%%%%%%%%%%%%%%%%%%%%%%%%%%%%%%%%%%%%%%%%%%%

%%%%%%%%%%%%%%%%%%%%%%%%%%%%%%%%%%%%%%%%%%%%%%%%%%%%%%%%%%%%%%%%%%%%%%
%% Table: Summary strategies
\begin{table*}[ht]  
\centering
\ra{1.3} 
\caption{Comparison of strategies. Median AUC and PR-AUC of the 5 runs, except for Wu~\etal~\cite{Wu2019Breast}. The performance of the first row evaluates the pretrained weights from \cite{Wu2019Breast} without further training on our 3 datasets. The second group of methods evaluates the cross-site performances (train in one/test in other). The third group shows the performance of federated approaches and the last is a non-privacy preserving oracle mix approach. The highlighted values in bold correspond to the best federated method.}
\begin{tabular}{@{}l|cc|cc|cc|cc@{}} \toprule
\multirow{2}{*}{} &
      \multicolumn{2}{c|}{\textbf{Hologic}} &
      \multicolumn{2}{c|}{\textbf{Siemens}} &
      \multicolumn{2}{c|}{\textbf{GE}} &
      \multicolumn{2}{c}{\textbf{AVG}} \\
      & AUC & PR-AUC & AUC & PR-AUC & AUC & PR-AUC & AUC & PR-AUC  \\
\midrule
Wu~\etal~\cite{Wu2019Breast} & 0.65 & 0.69 & 0.67 & 0.75 & 0.79 & 0.78 & 0.70 & 0.73 \\
\midrule
trHologic & - & - & 0.67 & 0.74 & 0.73 & 0.74 & - & -  \\
trSiemens & 0.59 & 0.63 & - & - & 0.65 & 0.67 & - & -  \\
trGE & 0.64 & 0.66 & 0.72 & 0.79 & - & - & - & -  \\
Single & 0.83 & 0.84 & 0.83 & 0.84 & 0.85 & 0.83 & - & -  \\ \midrule
Fed & 0.78 & 0.78 & 0.65 & 0.74 & 0.83 & \textbf{0.83} & 0.75 & 0.77 \\
Fed-CL & 0.80 & 0.80 & 0.63 & 0.72 & 0.81 & 0.81 & 0.75 & 0.78 \\
Fed-Align & 0.79 & 0.78 & 0.69 & \textbf{0.79} & \textbf{0.85} & \textbf{0.83} & 0.78 & 0.80 \\
Fed-Align-CL & \textbf{0.84} & \textbf{0.84} & \textbf{0.70} & \textbf{0.79} & 0.83 & 0.82 & \textbf{0.79} & \textbf{0.82}  \\
\midrule
Mix & 0.83 & 0.84 & 0.86 & 0.83 & 0.82 & 0.88 & 0.84 & 0.85 \\
\bottomrule
\end{tabular}
\label{table:comparison}
\end{table*}
%%%%%%%%%%%%%%%%%%%%%%%%%%%%%%%%%%%%%%%%%%%%%%%%%%%%%%%%%%%%%%%%%%%%%%

\section{Results} \label{sec:results}
\paragraph{\textbf{Initialization of local models}} 
First of all, we investigate the classification performance of the FL method with different pretraining strategies. The first case, \textit{Local model}, corresponds to pretraining each model with their own private data. The second case, \textit{Scratch}, refers to a random initialization of the local model weights. The third case, \textit{DDSM}, corresponds to pretraining the models on the CBIS-DDSM dataset \cite{Lee2017CBISDDSM}. The last case corresponds to initializing the model with the publicly shared weights from Wu~\etal~\cite{Wu2019Breast}.

In Table~\ref{table:initialization}, the AUC for the different initialization strategies is reported. In the columns we report the AUC for each site, and then averaged over sites (last column). We found that the best approach was using the pretrained weights from Wu~\etal~\cite{Wu2019Breast}. This behaviour is expected because their model was trained with a very-large private dataset. Moreover, the model in \cite{Wu2019Breast} was already pretrained with the ImageNet~\cite{Krizhevsky2012imagenet} dataset. Interestingly as well, classification results were better when all local models were initialized either randomly or pretrained on a single dataset (DDSM) than when each of them was pretrained on a small private dataset. Although DDSM dataset is large, it is formed by screen film mammography instead of FFDM, which explains the difference to Wu~\etal's weights. 

\paragraph{\textbf{Comparison with different strategies}}
To demonstrate the performance of our proposed method (\textit{Fed-Align-CL}), we compared it with three non-federated strategies and with two federated strategies. The non-federated strategies consist of: (i) training and testing within a single site (\textit{Single}); (ii) training using one site and testing on another site (\textit{Cross}); and (iii) collecting multi-site data together for both training and testing (\textit{Mix}). The later does not preserve data privacy since this model requires access to all training images and their respective classification labels. In \textit{Cross}, we denote the site used for training as `tr$<$site$>$`. Also, we ignore the performance of the site used for training in this row, and report it in the row `Single'. The federated strategies consist of training a client-server-based FL method with: (iv) \texttt{FedAvg}~\cite{McMahan2017FedAvg}, and (v) federated adversarial learning~\cite{Peng2019FedDA, Li2020FLDA}. We also performed an ablation study to verify the individual contributions of the domain alignment and the curriculum scheduling. Therefore, we included in our comparison: \textit{Fed}, \textit{Fed-Align} and \textit{Fed-CL}.

Classification metrics for breast malignancy classification are reported in Table~\ref{table:comparison}. In the first row, we include the performance of Wu's model \cite{Wu2019Breast} without further training. In the first place, we present the results of the non-federated methods. First, as expected, we find that the \textit{Cross} models do not generalize well across manufacturers. Such inconsistent performance is a known problem of deep learning methods for mammography classification~\cite{Wang2020Inconsistent}. Second, the individual models (\textit{Single}) achieve an average AUC of 0.83 for the three sites. When comparing our performance to other works on the publicly available INBreast dataset (Siemens), we achieve an AUC comparable to \cite{Wu2019Breast}, but lower than \cite{Ribli2018, Shen2019} with an AUC of 0.95. However, the later two works rely on region-wise ground truth: the first leveraging ROI localization and the second one using patch pretraining. In contrast, our models only rely on the full mammograms and their corresponding classification label. As expected, the best performing model is \textit{Mix}, which is trained with mammography images and their corresponding annotations from all sites, thus, not preserving privacy.

\begin{figure*}[]
    \centering
    \includegraphics[width=0.9 \textwidth]{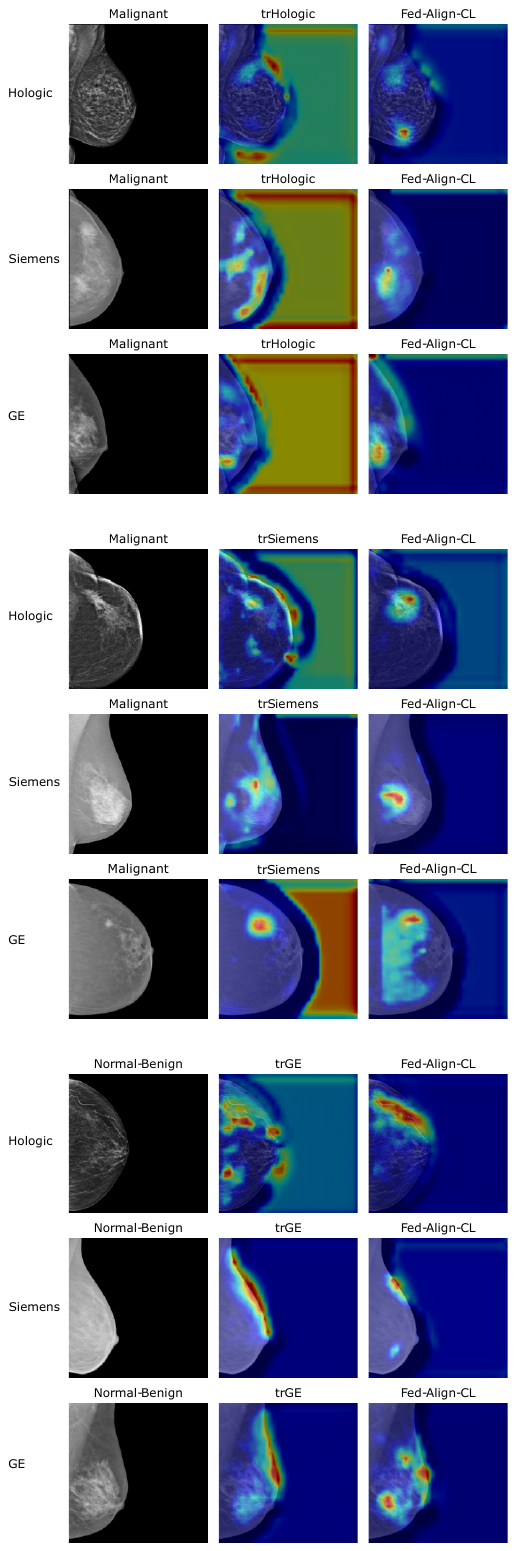}%
    \caption{Grad-CAM visualization of examples from single site and our federated approach. }%
    \label{fig:generalization}%
\end{figure*}

In the second place, we compare the federated approaches. First, we find that the \textit{Fed-CL} method improves on average the PR-AUC with respect to \textit{Fed}. However, the performance for the different domains of these two methods can be uneven. The \textit{Fed-Align} approach helps to learn domain-invariant features that are beneficial for the classification task. Finally, we can see that our proposed \textit{Fed-Align-CL} achieves on average the highest AUC and PR-AUC. We also find a consistent improvement with our proposed method when the models are trained from scratch. However, the classification metrics were better with the pretrained weights~\cite{Wu2019Breast}, as discussed in the previous experiment. We applied one-way ANOVA followed by Bonferroni's post-hoc t-test comparisons tests to evaluate for differences in the performance of the federated approaches and reported the p-values in Table 5 (Suppl. Material). Our analysis shows that the classification improvement of our method is statistically significant. Some qualitative results using Gradient-weighted Class Activation Mappings (Grad-CAMs)~\cite{Selvaraju2017GradCAM} show the generalisation effect of the federated approaches in Fig.~\ref{fig:generalization} (Suppl. Material).

\begin{figure*}[t]
    \centering
    \includegraphics[width=1.0
    \textwidth]{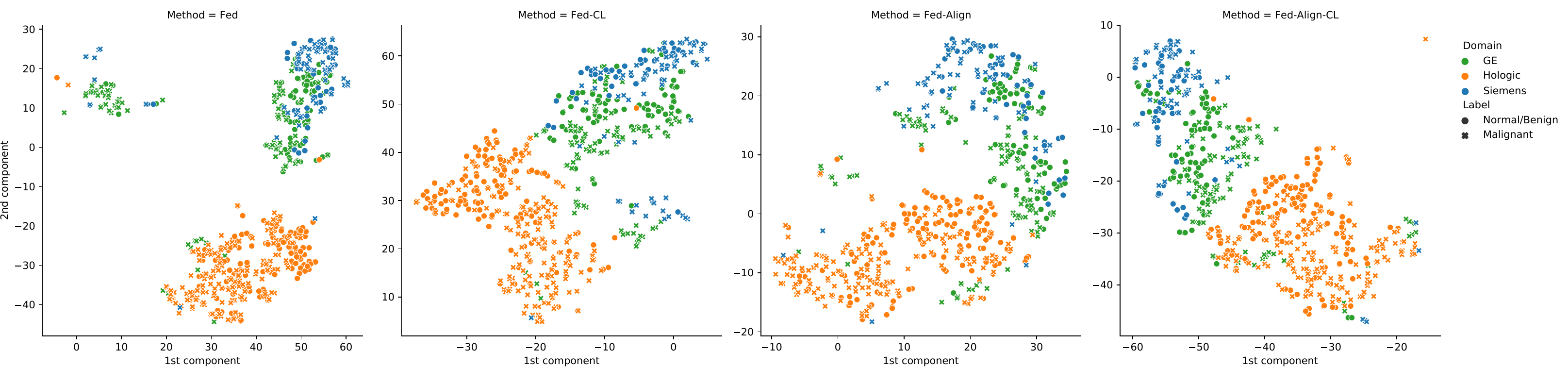}%
    \caption{t-SNE visualization of the latent space obtained by Fed, Fed-CL, Fed-Align and Fed-Align-CL in that order. The circles represent normal and benign samples, and the crosses malignant cases. Each color represents a domain. }%
    \label{fig:tSNE}%
\end{figure*}

\paragraph{\textbf{Alignment of features in latent space}} In order to visualize the effect of the domain adaptation and curriculum scheduling techniques, we show in Fig.~\ref{fig:tSNE} the two-dimensional t-SNE~\cite{VanMaaten2008tSNE} projection of the embedded latent space. First, we can see that the features learned by \textit{Fed} are better clusterized according to the input domain, \ie \textit{Fed} learns domain-variant features. Second, the combination of FL with CL results in samples more spread along the manifold, although still dependent on the domain. In the plots that correspond to the models that perform DA, we can see that the input images are better clusterized according to the label instead of the domain. We find that the domain alignment is particularly helpful for Siemens. Fig.~\ref{fig:tSNEpenultimate} in Supplementary Material shows a similar behaviour for the penultimate classification layer.

\begin{figure*}[h]
    \centering
    \includegraphics[width=0.6 \textwidth]{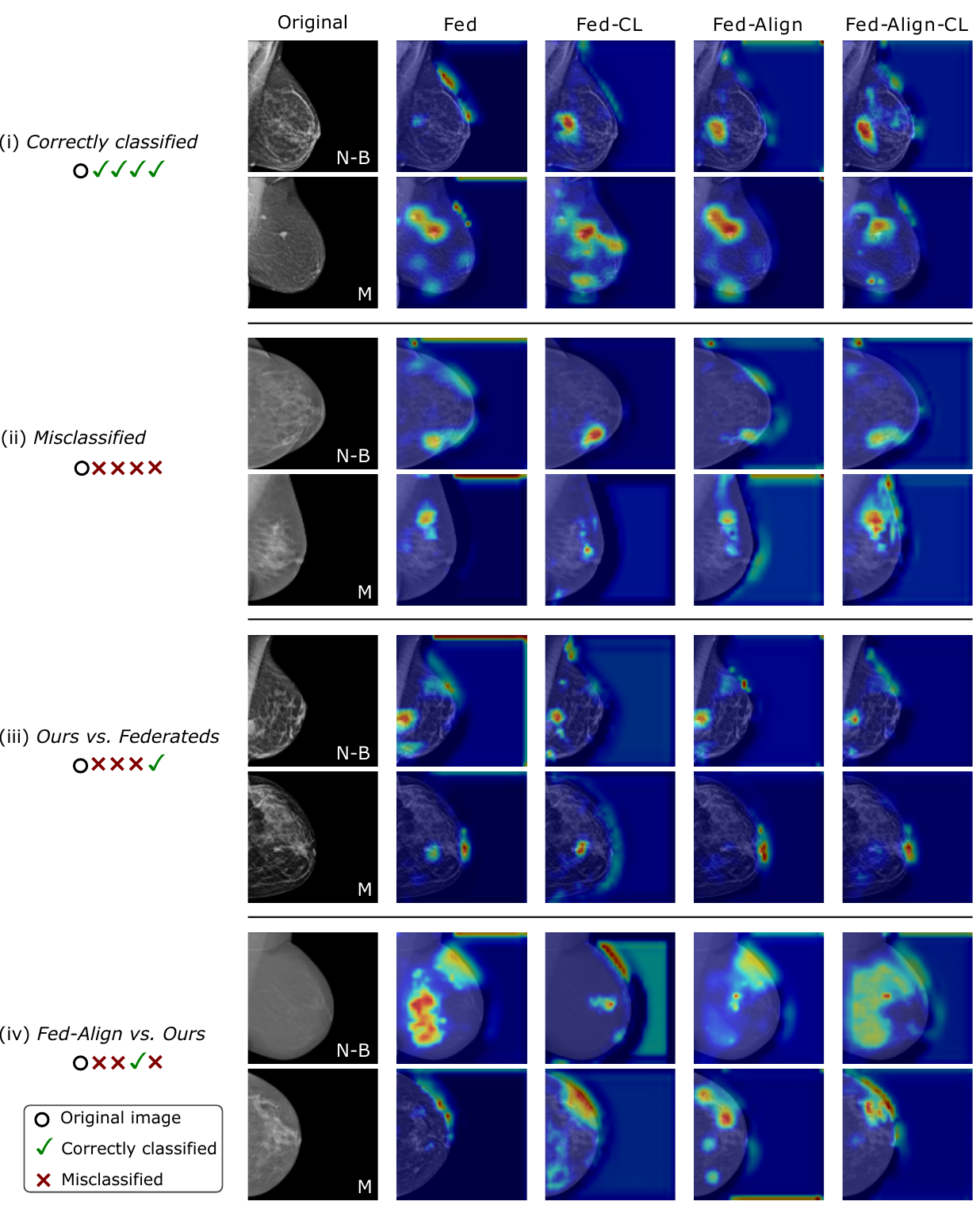}%
    \caption{Grad-CAM visualization of examples from Normal-Benign (N-B) and Malignant (M) classes. The first column depicts the original image, and the subsequent columns each of the models. Every pair of rows is a different scenario: (i) \textit{Correctly classified}, (ii) \textit{Misclassified}, (iii) \textit{Ours. \vs Federateds}, (iv) \textit{Fed-Align \vs Ours}.}%
    \label{fig:gradcam}%
\end{figure*}

\paragraph{\textbf{Examination of misclassified samples}} First, we analyzed the misclassified mammograms by the different models. We noted that the proposed method (Fed-Align-CL) outperforms the compared approaches in uncertain and difficult cases. That is, our method allows to avoid the misclassification of benign samples containing benign calcifications and benign distortions (\eg{} surgery remainings). Moreover, some of the barely visible masses are correctly captured as malignant by our method. Second, to better understand the effect of different strategies, we compared their resulting Grad-CAMs in several scenarios. We present a comparison of the strategies based on their classification performance and considering four scenarios: (i) \textit{Correctly classified} images by all federated models; (ii) \textit{Misclassified} images by all federated models, (iii) \textit{Ours \vs Federateds}: correctly classified images by our model (Fed-Align-CL) and misclassified by the other federated models; (iv) \textit{Fed-Align \vs Ours}: correctly classified images by Fed-Align and misclassified by our model (Fed-Align-CL). We show examples of the different scenarios for Normal-Benign and Malignant classes in Fig.~\ref{fig:gradcam}. We found that our proposed method correctly classifies a larger percentage of images from the Malignant class than other methods. Our method (Fed-Align-CL) reduces activations on the boundary or outside the breast and better focuses on lesions.

%%%%%%%%%%%%%%%%%%%%%%%%%%%%%%%%%%%%%%%%%%%%%%%%%%%%% Figure 
%%%%%%%%%%%%%%%%%%%%%%%%%%%%%%%%%%%%%%%%%%%%%%%%%%%%%%%%%%%%%%%%%%%%%%

%%%%%%%%%%%%%%%%%%%%%%%%%%%%%%%%%%%%%%%%%%%%%%%%%%%%% Figure 
%%%%%%%%%%%%%%%%%%%%%%%%%%%%%%%%%%%%%%%%%%%%%%%%%%%%%%%%%%%%%%%%%%%%%%

\section{Discussion} \label{sec:discussion} 
FL is a potential solution for the future of digital health \cite{Rieke2020FutureFL}, especially for classification tasks without access to sufficient data. FL allows for collaboratively training a model without sharing private data from different sites. A challenging aspect of sharing data within FL is that related to legal regulations and ethics. Privacy and data-protection need to be taken carefully into account. High-quality anonymization from a mammography or electronic record has to be guaranteed and in certain regions GDPR\footnote{GDPR: EU/UK General Data Protection Regulation}~\cite{Horvitz2015GDPR} or HIPAA\footnote{HIPAA: Health Insurance Portability and Accountability Act}~\cite{Annas2003HIPAA} compliant. Privacy-preserving techniques for FL provide a trade-off between model performance and reidentification. However, remaining data elements may allow for patient reidentification \cite{Rocher2019PatientReID}. Unless the anonymization process destroys the data fidelity, patient reidentification or information leakage cannot be discarded.

Another challenging aspect of FL is training a model with mixed heterogeneous data, \ie images obtained from different system vendors or acquisition protocols. In this work, we have investigated and confirmed the negative effect of domain shift for malignancy classification with multi-site mammograms. Models trained on single-vendor images do not generalize adequately to other sites. While some sites preserve the performance on the local test set after collaborative training, others trade a significant gain in generalization for a local performance drop. Although a drop can initially seem unacceptable, it is to be weighted with the improved robustness of the collaborative model to unseen data. By focusing on correcting local errors induced by the latest global update, our curriculum is a step towards closing the local-global performance gap. However, there is still some room for improvement when dealing with small datasets, such as Siemens. With the same collaborative training, there is no onsite performance drop for the GE and Hologic sites, which benefit from a significant generalization improvement at no cost. Regarding the federated models on the GE site, the domain adaptation component has a higher effect than the curriculum. We explain this behavior by the similarity between the data distribution of the GE site and the dataset used as pretraining for the initial local models. In this sense, domain alignment keeps the performance from dropping, but there are fewer corrections for the curriculum to account for after the update.
Finally, apart from the quantitative performance evaluation, we used Grad-CAM to produce qualitative results allowing a deeper discussion of the effects of CL and DA on the feature learning. We are aware of the subjective nature of interpreting such images \cite{Zhang2022TrustworthinessMaps}. For a more quantitative explainability it would be interesting to explore \cite{VANDERVELDEN2022XAI}.

The computational cost of analyzing high-resolution images imposes some resource limitations on our federated scenario. Indeed, the small size of the lesion requires high-resolution images showing tissue details to provide a precise diagnosis. An auxiliary localization model focusing on potentially positive regions could be beneficial and complementary to our proposed method. However, the level of annotation required would be much higher as standard clinical protocols do not include the lesion delineation. Pretrained detectors on natural images for region proposals, such as R-CNN~\cite{Girshick2014RCNN} or YOLO~\cite{Redmon2016YOLO} and models derived from them, are likely ineffective given their low resolution. To better focus on the breast lesions, one could also investigate the use of attention models, such as Transformers~\cite{Dosovitskiy2020Transformers} or the most recent ConvNext~\cite{Liu2022ConvNeXt}. This type of models could be leveraged either on low-resolution images or scaling up the resolution of the attention models.

In this work, we have investigated the use of CL to boost the alignment between domain pairs and improve the overall classification of breast cancer. In particular, our memory-aware curriculum is implemented with a data scheduler that arranges the order of the training samples to improve model consistency in the federated setting. This order is defined with a scoring function that prioritizes training samples that have been forgotten after the deployment of the global model. Our method, inspired by the semi-supervised learning methods of temporal ensembling~\cite{Laine2017TemporalEnsembling} and consistency training~\cite{Xie2020ConsistencyTraining}, aims to deliver consistent predictions. We believe that enforcing consistent predictions may act as a regularizer. Furthermore, our proposed method can be effective to deal with performance inconsistency across data sets and models~\cite{Wang2020Inconsistent}. We believe further research will follow on the use of CL in combination with FL and DA. We envision three approaches which we discuss next: prioritizing training (source) samples for better classification; smartly weighting the aggregation of the local models; and improving alignment between domains pairs. For instance, one could leverage other schemes motivated by boosting~\cite{Freund1999Boosting}, that have been studied in active learning or CL but have not yet been used for federated settings. These schemes can be designed to prioritize the (source) samples via a data scheduler. 
Regarding the local model aggregation, one could deploy a CL-based adaptive weighting for clients based on a dynamic scoring function taking into account meta-information~\cite{Yeganeh2020IDA}, and in this way, help to cope with unbalanced and non-IID data. Monitoring schemes to infer the composition of training data for each FL round, and designing new losses functions could help to mitigate the impact of imbalance \cite{Wang2021FLIMbalance, Shen2021Agnostic}. Finally, to improve alignment, scoring functions could rely on computing the distance between (noisy) latent representations of the source and the remaining domains to weigh each local model contribution.

\section{Conclusions}
\label{sec:conclusions}
In this work, we have designed and integrated a CL strategy in a federated adversarial learning setting for the classification of breast cancer. We have learned a collaborative decentralized model with three clinical datasets from different vendors. We have shown that, by monitoring the local and global classification predictions, we can schedule the training samples to boost the alignment between domain pairs and improve the classification performance.

\section*{Acknowledgements}
This project has received funding from the European Union’s Horizon 2020 research and innovation programme under the Marie Sk\l{}odowska-Curie grant agreement No. 713673 and by the Spanish Ministry of Economy [MDM-2015-0502]. A. \JimSan{} has received financial support through the ``la Caixa'' Foundation (ID Q5850017D), fellowship code: LCF/BQ/IN17/11620013. D. Mateus has received funding from Nantes M\'etropole and the European Regional Development, Pays de la Loire, under the Connect Talent scheme. Authors thank Nvidia for the donation of a GPU.

\bibliographystyle{IEEEtran}
\bibliography{IEEEabrv,refs}

\appendices

\clearpage
\newpage

\section*{Supplementary Material}

\paragraph{\textbf{Algorithm~\ref{alg:framework}}} presents the pseudo-code for our novel memory-aware curriculum federated learning.

\paragraph{\textbf{Architecture of the models}}
We provide the detailed model architecture in Table~\ref{table:architecture}. We denote convolutional layers as \textit{Conv}, max pooling layers as \textit{MaxPool}, fully connected layers as \textit{FC}, batch normalization layers as \textit{BN}, ReLu layers as \textit{ReLu}, dropout layers as \textit{Dropout} and sigmoid layers as \textit{Sigmoid}. For FC layers, the values in brackets represent the input and output dimensions. For Conv layers, we provide in this order: the input and output feature maps, the kernel size, the stride and the padding. For MaxPool layers, we provide in this order: kernel size, stride, padding and dilation. For dropout layers (Dropout), we provide the probability of an element to be zeroed. To define the feature extractor, we define a \textit{Block} which consists of a series of layers and specify for the convolutional layers: the input and output feature maps, the kernel size, the stride for the first convolution \textit{$s_1$}, the stride for the second convolution \textit{$s_2$}, and the padding. In particular Block consists of: ReLu, BN, Conv(), BN and Conv().

\paragraph{\textbf{Statistical significance}} In Table~\ref{table:pvalues}, we applied one-way ANOVA followed by Bonferroni's post-hoc t-test comparisons tests to evaluate for differences in the performance of the federated approaches and reported the p-values.

\begin{figure*}[]
    \centering
    \includegraphics[width=1.0 \textwidth]{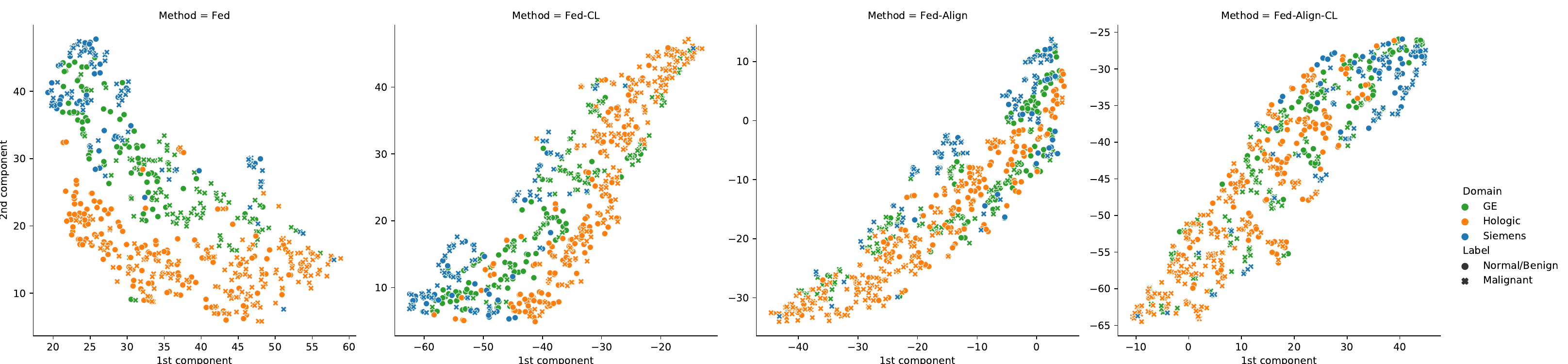}%
    \caption{t-SNE visualization of the penultimate classification layer obtained by Fed, Fed-CL, Fed-Align and Fed-Align-CL in that order. The circles represent normal and benign samples, and the crosses malignant cases. Each color represents a domain. }%
    \label{fig:tSNEpenultimate}%
\end{figure*}

\paragraph{\textbf{t-SNE feature visualization}} Figure~\ref{fig:tSNEpenultimate} depicts the first two components after applying t-SNE to the penultimate classification layer of every federated method.

%%%%%%%%%%%%%%%%%%%%%%%%%%%%%%%%%%%%%%%%%%%%%%%%%%%%%%%%%%%%%%%%%%%%%%
%% Table: architectural details
\begin{table}[b]  % bht
\centering
\ra{1.5}
\caption{ResNet-22 architecture for breast cancer classification.}
\begin{tabular}{@{}l|c@{}} \toprule
    \textbf{Layer} & \textbf{Configuration}  \\ \midrule
    & \textbf{\textbf{F: Feature Extractor}} \\ \midrule
    1 & \multicolumn{1}{l}{Conv(1, 16, 3, 1, 1), MaxPool(3, 2, 0, 1)} \\
    2.1 & \multicolumn{1}{l}{Block(16, 16, 3, $s_1$=1, $s_2$=1, 1), Conv(16, 16, 1, 1)} \\ 
    2.2 & \multicolumn{1}{l}{Block(16, 32, 3, $s_1$=1, $s_2$=1, 1)} \\ 
    3.1 & \multicolumn{1}{l}{Block(16, 32, 3, $s_1$=2, $s_2$=1, 1), Conv(16, 32, 1, 2)} \\ 
    3.2 & \multicolumn{1}{l}{Block(16, 32, 3, $s_1$=1, $s_2$=1, 1)} \\ 
    4.1 & \multicolumn{1}{l}{Block(32, 64, 3, $s_1$=2, $s_2$=1, 1), Conv(32, 64, 1, 2)} \\ 
    4.2 & \multicolumn{1}{l}{Block(32, 64, 3, $s_1$=1, $s_2$=1, 1)} \\ 
    5.1 & \multicolumn{1}{l}{Block(64, 128, 3, $s_1$=2, $s_2$=1, 1), Conv(64, 128, 1, 2)} \\ 
    5.2 & \multicolumn{1}{l}{Block(64, 128, 3, $s_1$=1, $s_2$=1, 1)} \\ 
    6.1 & \multicolumn{1}{l}{Block(128, 256, 3, $s_1$=2, $s_2$=1, 1), Conv(128, 256, 1, 2)} \\ 
    6.2 & \multicolumn{1}{l}{Block(128, 256, 3, $s_1$=1, $s_2$=1, 1)} \\ 
    \midrule
    & \textbf{\textbf{Cls: Classifier}} \\ \midrule
    1 & \multicolumn{1}{l}{FC(256, 128), BN, ReLu, Dropout(0.5)} \\
    2 & \multicolumn{1}{l}{FC(128, 64), BN, ReLu, Dropout(0.5)} \\
    3 & \multicolumn{1}{l}{FC(64, 2), Sigmoid} \\
    \midrule
    & \textbf{\textbf{D: Domain Discriminator}} \\ \midrule
    1 & \multicolumn{1}{l}{FC(256, 4), ReLu} \\
    2 & \multicolumn{1}{l}{FC(4, 2), Sigmoid} \\
\bottomrule
\end{tabular}
\label{table:architecture}
\end{table}
%%%%%%%%%%%%%%%%%%%%%%%%%%%%%%%%%%%%%%%%%%%%%%%%%%%%%%%%%%%%%%%%%%%%%%

%%%%%%%%%%%%%%%%%%%%%%%%%%%%%%%%%%%%%%%%%%%%%%%%%%%%%%%%%%%%%%%%%%%%%%
%% Table: p-values comparison strategies
\begin{table*}[]
\centering
\ra{1.5} %The higher the better. 
\caption{P-values to test statistical significance of the 5 runs among the different methods.}
\begin{tabular}{l|c|c|c|c|c|c}
\toprule
\multirow{2}{*}{} & 
    \multicolumn{2}{c|}{Fed}& 
    \multicolumn{2}{c|}{Fed-CL} & 
    \multicolumn{2}{c}{Fed-Align} \\
    & AUC & PR-AUC & AUC & PR-AUC & AUC & PR-AUC \\
\hline
Fed-CL & $7.60E-01$ & $7.13E-01$ & - & - & - & - \\
Fed-Align & $2.34E-03$ & $1.51E-03$ & $1.55E-01$ & $1.82E-01$ & - & -\\
Fed-Align-CL & $1.46E-04$ & $9.82E-06$ & $5.75E-02$ & $3.29E-02$ & $5.65E-02$ & $1.42E-02$ \\
\bottomrule
 \end{tabular}
\label{table:pvalues}
\end{table*}

%%%%%%%%%%%%%%%%%%%%%%%%%%%%%%%%%%%%%%%%%%%%%%%%%%%%% Figure 
%%%%%%%%%%%%%%%%%%%%%%%%%%%%%%%%%%%%%%%%%%%%%%%%%%%%%%%%%%%%%%%%%%%%%%

%%%%%%%%%%%%%%%%%%%%%%%%%%%%%%%%%%%%%%%%%%%%%%%%%%%%% Figure 
%%%%%%%%%%%%%%%%%%%%%%%%%%%%%%%%%%%%%%%%%%%%%%%%%%%%%%%%%%%%%%%%%%%%%%

%%%%%%%%%%%%%%%%%%%%%%%%%%%%%%%%%%%%%%%%%%%%%%%%%%%%%%%%%%%%%%%%%%%%%%%%%%%%%%%%%%%%
%%%%%%%%%%%% Algorithm %%%%%%%%%%%%
\begin{algorithm*}[]
\SetAlgoLined
\SetKwInOut{Input}{input}
\SetKwInOut{Output}{output}
\Input{$N$ number of sites, $X=\{X_1,..., X_N\}$ mammograms, $Y=\{Y_1,..., Y_N\}$ classification labels, \\ $\{X_i^S, Y_i^S\}_ {i=1}^{N}$ source dataset, $\{X_j^T\}_ {j=1}^{N}$ target dataset, $m_n$ training size at site $n$, \\ $f_w = \{f_{w_1},...f_{w_N}\}$ local models, $Z(\cdot)$ noise generator for privacy-preserving, \\ $Q$ number of optimization iterations, $\tau$ communication pace, $E$ optimization epochs, $E_w$ warm-up epochs, \\
$\{\text{opt}_1(\cdot), ..., \text{opt}_N(\cdot)\}$ optimizers}
\Output{global model: $g_w$} 
\texttt{\\}
Initialize local models: $\{f_{w_0}, ..., f_{w_N}\} \xleftarrow[]{}$ Wu~\etal~\cite{Wu2019Breast} \\ %\algorithmiccomment{initialize local model} \\
\For{$e=1$ \KwTo E}{
    \uIf{$e > E_w$}{
    \For{$i = 1 $ \KwTo $N$ }{         
        \texttt{\\}
        \textbf{Memory-aware curriculum} \\
        \For{$k = 1 $ \KwTo $m_i$ }{
        Get the local $\hat{y}_{L,k}^{(e)}$ and global $\hat{y}_{G,k}^{(e)}$ classification predictions \;
        Compute the curriculum weights $\rho_{i,k}^{(e)}$ with Eq.~\eqref{eq:clweights} \; 
        %$\pi^{(e)}$ \algorithmiccomment{reordering function obtained by sampling with $\rho_k$} \\
        Obtain reordering function $\pi_i^{(e)}$ by sampling with $\rho_i^{(e)}$ \;
        Reorder training data: $\{X,Y\} \xrightarrow[\pi_i^{(e)}]{} \{X_{\pi_i^{(e)}}, Y_{\pi_i^{(e)}}\} $ \;
        }
    }
    }
    \Else{
    Random permutation $\pi_i^{(e)}$ \;
    }
    \texttt{\\}
    \For{$q = 1 $ \KwTo $Q$ }{
        %Initialize pace counter:  $t \xleftarrow[]{} 0$ \; \\
        %\algorithmiccomment{initialize pace counter} \\
        \For{$i = 1 $ \KwTo $N$ }{
        \texttt{\\}
        \textbf{Local classification} \\
        Get the \textbf{next} mini-batch from source site $i \, \{X_{\pi_i^{(e)}, b}^S, Y_{\pi_i^{(e)}, b}^S\}_{b=1}^{B_i*Q}$ \;
        Compute classification loss $\mathcal{L}_{Cls}(f_{w_i^{(q-1)}}(X^S_{\pi_i^{(e)},b},Y^S_{\pi_i^{(e)},b})))$ with Eq.~\eqref{eq:lossCls} \;
        % Update each local site's weights
        Update $w_{F_i}^{(q)}$, $w_{{Cls}_{i}}^{(q)} \xleftarrow[]{} opt_i(\mathcal{L}_{Cls})$ \; 
        %\algorithmiccomment{update $w_{F_i}^{(m)}$, $w_{Cls}_{i}^{(m)}$} \\
        \texttt{\\}
        \uIf{$e > E_w$}{
        \For{$j = 1 $ \KwTo $N$ \textbf{and} $j \neq i $}{
        \texttt{\\}
        \textbf{Domain alignment} \\
        Get the \textbf{next} mini-batch from target site $j \, \{X_{\pi_j^{(e)}, b}^T\}_{b=1}^{B_j*Q}$ \;
        Compute adversarial loss $\mathcal{L}_{D}$ with Eq.~\eqref{eq:lossD} \;
        Update $w_{{D_i}}^{(q)} \xleftarrow[]{} opt_i(\mathcal{L}_{D})$ \; 
        Compute feature extractor loss $\mathcal{L}_{F}$ with Eq.~\eqref{eq:lossF}  \;
        %\algorithmiccomment{update $\{ w_{G_i}^{(m)}, w_{G_j}^{(m)} \}$} \\
        Update $\{ w_{{F_i}}^{(q)},  w_{F_j}^{(q)} \} \xleftarrow[]{} opt_i(\mathcal{L}_{F})$ \; 
        }
        }
        }
        \texttt{\\}
        %$t \xleftarrow[]{} t + 1 $  \: \\
        %\algorithmiccomment{models communicate} \\
        \uIf{$q\,\%\,\tau = 0$ }{
        \texttt{\\}
        \textbf{Update global model} \\
        $\bar{w}^{(q)}_{F} \xleftarrow[]{} \dfrac{1}{N} \sum_{i=1}^{N}(w_{F_i}^{(q)} + Z(w_{F_i,}^{(q)})$ \; 
        $\bar{w}^{(q)}_{Cls} \xleftarrow[]{} \dfrac{1}{N} \sum_{i=1}^{N}(w_{{Cls}_i}^{(q)} + Z(w_{{Cls}_i}^{(q)})$ \;
        %\algorithmiccomment{update global model per $\tau$ steps} \\
        \textbf{Deploy weights to local model} \\
        \For{$i = 1 $ \KwTo $N$ }{
        ${w}^{(q)}_{F_i} \xleftarrow[]{} \bar{w}^{(q)}_{F}$ \;
        ${w}^{(q)}_{{Cls}_i} \xleftarrow[]{} \bar{w}^{(q)}_{Cls}$ \;
        }
      }
    }
}
\textbf{Return global model: } $g_w \xleftarrow[]{} ({w}^{(E)}_{F}, {w}^{(E)}_{Cls})$
\caption{Memory-aware Curriculum Federated Learning}
 \label{alg:framework}
\end{algorithm*}
%%%%%%%%%%%%%%%%%%%%%%%%%%%%%%%%%%%%
%%%%%%%%%%%%%%%%%%%%%%%%%%%%%%%%%%%%%%%%%%%%%%%%%%%%%%%%%%%%%%%%%%%%%%%%%%%%%%%%%%%%

\end{document}